\documentclass[runningheads]{llncs}
\usepackage{graphicx}

\usepackage{tikz}
\usepackage{comment}
\usepackage{amsmath,amssymb} 
\usepackage{color}
\usepackage{multirow}                           
\usepackage{booktabs}                           
\usepackage{colortbl}                           
\usepackage{siunitx}
\usepackage[noabbrev,capitalise]{cleveref}      
\usepackage{algpseudocode}                      
\usepackage{algorithm}                          
\usepackage{setspace}                           
\usepackage{caption}

\newcommand{\linenum}[1]{\small{#1}:}
\newcommand{\of}{\texttt{oflib}}
\newcommand{\Of}{\texttt{Oflib}}
\newcommand{\ofnp}{\texttt{oflibnumpy}}

\newcommand{\ofpt}{\texttt{oflibpytorch}}
\newcommand{\Ofpt}{\texttt{Oflibpytorch}}

\newcommand{\f}[2]{\mathcal{F}_{#1 \shortto #2}}
\newcommand{\fs}[2]{\mathcal{F}_{\underline{#1} \shortto #2}}
\newcommand{\ft}[2]{\mathcal{F}_{#1 \shortto \underline{#2}}}

\newcommand{\F}[2]{\mathbf{F}_{#1 \shortto #2}}
\newcommand{\Fs}[2]{\mathbf{F}_{\underline{#1} \shortto #2}}
\newcommand{\Ft}[2]{\mathbf{F}_{#1 \shortto \underline{#2}}}

\newcommand{\rf}[2]{\mathcal{F\,}^{#1}_{#2}}
\newcommand{\rftop}[2]{\mathcal{F\,}^{\underline{#1}}_{#2}}

\newcommand{\sign}[2]{\text{sign\,}^{#1}_{#2}}
\newcommand{\tref}[2]{{t\,}^{#1}_{#2}}

\newcommand{\E}[1]{$\mathcal{E}_{#1}$}

\usepackage{stmaryrd}
\usepackage{trimclip}
\makeatletter
\DeclareRobustCommand{\shortto}{%
  \mathrel{\mathpalette\short@to\relax}%
}
\newcommand{\short@to}[2]{%
  \mkern2mu
  \clipbox{{.45\width} 0 0 0}{$\m@th#1\vphantom{+}{\shortrightarrow}$}%
  }
\makeatother

\usepackage[accsupp]{axessibility}  

\begin{document}
\pagestyle{headings}
\mainmatter
\def\ECCVSubNumber{4}  

\title{Oflib: Facilitating Operations with and on Optical Flow Fields in Python} 

%
\author{Claudio S. Ravasio\inst{1,2}\orcidID{0000-0002-6453-5376} \and
Lyndon Da Cruz\inst{3}\orcidID{0000-0002-7695-6354} \and
Christos Bergeles\inst{2}\orcidID{0000-0002-9152-3194}}
\authorrunning{C. S. Ravasio et al.}
%
\institute{
    University College London (UCL), United Kingdom \and
    King's College London (KCL), United Kingdom \and
    Moorfields Eye Hospital, London, United Kingdom
}
\maketitle

\begin{abstract}
We present a robust theoretical framework for the characterisation and manipulation of optical flow, i.e $2$D vector fields, in the context of their use in motion estimation algorithms and beyond. The definition of two frames of reference guides the mathematical derivation of flow field application, inversion, evaluation, and composition operations. This structured approach is then used as the foundation for an implementation in Python $3$, with the fully differentiable PyTorch version \ofpt{} supporting back-propagation as required for deep learning. We verify the flow composition method empirically and provide a working example for its application to optical flow ground truth in synthetic training data creation. All code is publicly available.

\end{abstract}

\section{Introduction}

Optical flow as an expression of motion encoding and feature correspondence is one of the oldest tasks in computer vision, with seminal works such as by Lucas, Kanade et al \cite{lucas1981iterative} dating back to the early 80s. After decades of advances using variational methods, the extremely successful convolutional neural network based FlowNet method in $2015$ \cite{dosovitskiy_2015_flownet} heralded the arrival of well-performing and efficient end-to-end deep learning methods, usually implemented in Python. This has quickly become the dominant approach, with performance continuously improving and ever more complex benchmarks being proposed, such as MPI-Sintel, KITTI, or FlyingThings3D \cite{sintel_movie_2010,menze_2018_kitti,mayer_2016_large}.

In this context, handling optical flow easily and efficiently is of increasing importance. Many algorithms or their training protocols involve operations with or on optical flow fields, such as the creation of complex synthetic data \cite{ravasio_2020_learned} or working with ``cues'' calculated from bidirectional flow as proposed by Hofinger et al \cite{hofinger_2020_improving}. Implementing this from scratch can be laborious and error-prone. While there are a great number of publicly available algorithm implementations as well as methods in Python libraries for the estimation of optical flow fields \cite{opencv,farneback,liu}, no such wealth of resources exists for their further manipulation. An extensive search brought only little Python code to light, all being either algorithm-dependent or severely limited in their scope. Flow visualisation is an important topic, and the toolboxes \verb|flowvid| \cite{flowvid} as well as \verb|flow-vis| \cite{flow-vis} have some interesting capabilities in this regard. In addition to that, there are packages such as \verb|flowpy| \cite{flowpy} which also allow for basic flow warping, and add some utilities with a narrow focus on specific tasks such as reading and writing flows.

The aim of \of{} on the other hand is to offer a structured approach to the concept of flow fields, guided by a framework derived from first principles, and to provide all methods necessary to perform operations within a reasonable scope. This involves taking into account the two possible frames of reference for the flow vectors, as well as tracking undefined areas in outputs. The rigorous method ensures the mathematically correct implementation of a wide range of flow operations, including more complex functions -- such as flow composition -- not found in any of the previously listed python packages. Full interoperability with any Python code using NumPy \cite{numpy} or PyTorch \cite{pytorch} lends \of{} a high potential for reuse by the larger research community. The option to perform operations batched and on a GPU in particular yields significant speedups, while differentiability in the context of the Pytorch \verb|autograd| module allows for the use in any deep learning algorithm relying on back-propagation for optimisation.
\section{Theory}\label{theory}

The theoretical framework underpinning \of{} is derived from first principles to ensure a coherent and rigorous approach. This section will first address the two possible reference frames for optical flow fields and then present the theoretical basis for the main functionality provided by \of{}, focusing on the concrete operations needed to eventually translate the mathematical definition into code.

\subsection{Optical Flow Definition}
\label{theory:def}

\begin{figure}
    \centering
    \includegraphics[width=0.7\textwidth]{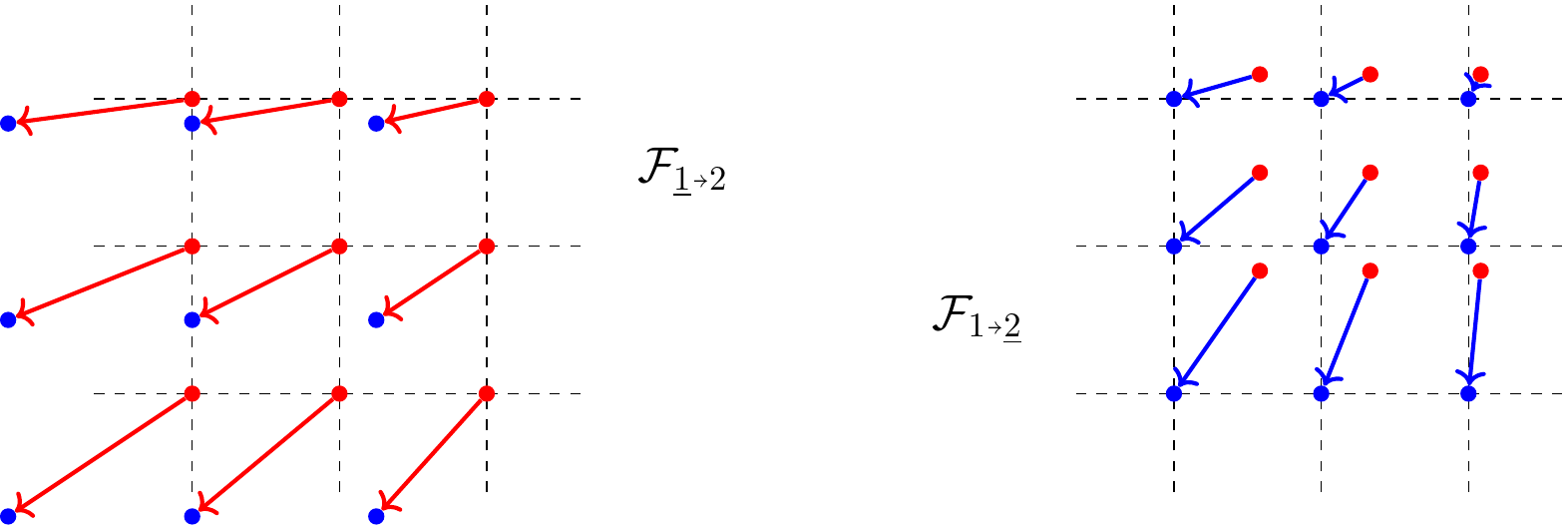}
    \caption{Two frames of reference, points at time $t_1$ in red, at $t_2$ in blue. \textbf{Left:} ``source'' means all pixels at time $t_1$, i.e. coordinates on a discrete grid $\mathbf{G}$, are mapped to a new location at time $t_2$.  \textbf{Right:} ``target'' means all pixels on this grid $\mathbf{G}$ at time $t_2$ are matched with a different previous location at time $t_1$.}
    \label{fig:flow_reference}
\end{figure}

An optical flow field is defined as a spatial mapping of coordinates at time $t_1$ to coordinates at time $t_2$:
\begin{equation}\label{eq:flow_def_gen}
    \mathcal{F}_{1 \shortto 2}:= \mathbf{X}_1 \mapsto \mathbf{X}_2; \quad \F{1}{2} = \mathbf{X}_{2} - \mathbf{X}_{1}
\end{equation}

\noindent where $\mathbf{X}_t$ corresponds to the set of continuous feature coordinates $\mathbf{x}$ being mapped at time $t$, and $\F{1}{2}$ is the resulting array of flow vectors between the feature sets at times $t_1$ and $t_2$. In the context of image sequences, this is equivalent to creating a mapping between the image feature coordinates $i_{\mathbf{X}_t}$ in the frame at time $t_1$ to coordinates in the frame at time $t_2$. We distinguish between two possible frames of reference, ``source'' and ``target''\footnote{The terms ``forward'' and ``backward'' flow often used in literature are avoided here, as it can lead to confusion e.g. in the context of reverse ``backward'' flows.}, illustrated in \cref{fig:flow_reference}:

\begin{itemize}
    \item \textbf{Source, ``s'':} In this case, coordinates on a discretised regular grid at time $t_1$, termed the ``source domain'', are mapped to coordinates in continuous space at time $t_2$. Applied to images, this indicates each pixel in the first image is matched with some position in the second image - but not every pixel at time $t_2$ has a known source correspondence at time $t_1$.
    \item\textbf{Target, ``t'':} The second option means that for each coordinate on a discretised regular grid at time $t_2$, or the ``target domain'', there is a mapping to a coordinate in continuous space at time $t_1$. Each pixel in the second image is matched with some position in the first image - but not every pixel at time $t_1$ has a known target correspondence at time $t_2$.
\end{itemize}

\Cref{eq:flow_def_gen} can therefore be extended as follows:
\begin{equation}\label{eq:flow_def}
\begin{split}
    \text{``Source'' reference: } \fs{1}{2} := \mathbf{G}_1 \mapsto \mathbf{X}_2; \quad \Fs{1}{2} = \mathbf{X}_2 - \mathbf{G}_1 \\
    \text{``Target'' reference: } \ft{1}{2} := \mathbf{X}_1 \mapsto \mathbf{G}_2; \quad \Ft{1}{2} = \mathbf{G}_2 - \mathbf{X}_1 \\
\end{split}
\end{equation}

\noindent where the underlined number in $\f{1}{2}$ indicates whether the source or the target of the mapping is on a discretised regular grid $\mathbf{G}$ in $2$D space, spanning the pixel range from $0$ to $H-1$ vertically and $0$ to $W-1$ horizontally, where $H$ and $W$ are the flow field height and width, respectively.

Note that while $\fs{1}{2} \neq -\fs{2}{1}$, as the inverse mapping of $\mathbf{G}_1 \mapsto \mathbf{X}_2$ is not $\mathbf{G}_2 \mapsto \mathbf{X}_1$, the following relationships do hold true:
\begin{equation}\label{eq:flow_inv_ref_switch}
\begin{split}
    \fs{1}{2} = \left.\ft{2}{1}\right\vert^\text{inv} \quad \Rightarrow \quad \mathbf{G}_1 \mapsto \mathbf{X}_2 = (\mathbf{X}_2 \mapsto \mathbf{G}_1)^\text{inv} \quad \Rightarrow \quad \Fs{1}{2} = -\Ft{2}{1}\\
    \ft{1}{2} = \left.\fs{2}{1}\right\vert^\text{inv} \quad \Rightarrow \quad \mathbf{X}_1 \mapsto \mathbf{G}_2 = (\mathbf{G}_2 \mapsto \mathbf{X}_1)^\text{inv} \quad \Rightarrow \quad \Ft{1}{2} = -\Fs{2}{1}
\end{split}
\end{equation}

\subsection{Flow Application}
\label{theory:apply}

Given data on the spatial grid $\mathbf{G}$ at time $t_1$ such as an image $i_{\mathbf{G}_1}$, an optical flow field $\mathcal{F}_{1 \shortto 2}$ on the same grid $\mathbf{G}$ can be applied to it to calculate the warped data $i'$ on $\mathbf{G}$ at time $t_2$, so that $\mathcal{F}_{1 \shortto 2}\{i_{\mathbf{G}_1}\} := i_{\mathbf{G}_1} \mapsto i'_{\mathbf{G}_2}$. The operation $\f{1}{2}\{\cdot\}$ is applied to a data array of shape $H \times W \times C$, with $C$ as the channel dimension containing data vectors. The output is a data array of the same shape. Keeping the two different frames of reference in \cref{eq:flow_def} in consideration:
\begin{equation}\label{eq:apply}
\begin{split}
    \fs{1}{2}\{i_{\mathbf{G}_1}\} &:= i_{\mathbf{G}_1} \mapsto i_{\mathbf{X}_2} \ \: \oplus \ \:
    \mathcal{I}\{i_{\mathbf{X}_2} \shortto i_{\mathbf{G}_2}\} \\
    \ft{1}{2}\{i_{\mathbf{G}_1}\} &:= \mathcal{I}\{i_{\mathbf{G}_1} \shortto i_{\mathbf{X}_1}\} \ \: \oplus \ \: i_{\mathbf{X}_1} \mapsto i_{\mathbf{G}_2}
\end{split}
\end{equation}

\noindent where $\oplus$ indicates sequential operations, and $\mathcal{I}\{i_\mathbf{X} \shortto i_\mathbf{Y}\} = \mathcal{I}\{i;\:\mathbf{X} \shortto \mathbf{Y}\}$ is the interpolation operation of data $i$ from coordinates $\mathbf{X}$ to $\mathbf{Y}$. It follows:
\begin{equation}\label{eq:apply_op}
\begin{split}
    i' = \fs{1}{2}\{i\} &= \mathcal{I}\{i;\:\mathbf{X}_2 \shortto \mathbf{G}\} = \mathcal{I}\{i;\:(\mathbf{G} + \Fs{1}{2}) \shortto \mathbf{G}\} \\
    i' = \ft{1}{2}\{i\} &= \mathcal{I}\{i;\:\mathbf{G} \shortto \mathbf{X}_1\} = \mathcal{I}\{i;\:\mathbf{G} \shortto (\mathbf{G} - \Ft{1}{2})\}
\end{split}
\end{equation}

As warped data consists of linear combinations of the initial data, $\mathcal{F}\{\cdot\}$ on data arrays $i$ and $j$ is a distributive operation, i.e. $\mathcal{F}\{i + j\} = \mathcal{F}\{i\} + \mathcal{F}\{j\}$.

\subsection{Point Tracking}
\label{theory:track}

Tracking points from $t_1$ to $t_2$ is conceptually closely related to applying an optical field to some input as detailed in \cref{theory:apply}. It can be defined as $\mathcal{F}_{1 \shortto 2}\langle{\mathbf{P}_1}\rangle := \mathbf{P}_1 \mapsto \mathbf{P}_2$, where $\mathbf{P}_t$ is a set of continuous point coordinates $\{\mathbf{p}_1, ..., \mathbf{p}_N\}$ at time $t$, located within the $2$D space spanned by the discretised regular grid $\mathbf{G}$ on which the optical flow field is defined. $\f{1}{2}\langle\rangle$ takes an array of shape $N \times 2$ containing point coordinates and outputs an array of the same shape $N \times 2$.

$\mathbf{P}_2$ is calculated by adding the flow vectors at $\mathbf{P}_1$ to $\mathbf{P}_1$. Given the two flow field references introduced in \cref{theory:def}, we obtain the following two equations:
\begin{equation}\label{eq:track}
\begin{split}
    \mathbf{P}_2 = \fs{1}{2}\langle{\mathbf{P}_1}\rangle &= \mathbf{P}_1 + \mathcal{I}\{\Fs{1}{2};\:\mathbf{G} \shortto \mathbf{P}_1\} \\
    \mathbf{P}_2 = \ft{1}{2}\langle{\mathbf{P}_1}\rangle &= \mathbf{P}_1 + \mathcal{I}\{\Ft{1}{2};\:(\mathbf{G} - \Ft{1}{2}) \shortto \mathbf{P}_1\} \\
\end{split}
\end{equation}

\noindent where $\mathcal{I}\{\mathbf{F};\:\mathbf{X} \shortto \mathbf{Y}\}$ is the interpolation of $\mathbf{F}$ from coordinates $\mathbf{X}$ to $\mathbf{Y}$.

\subsection{Reference Switch}
\label{theory:switch}

For a ``source'' reference flow field changing the reference frame means interpolating the flow vectors from their end points $\mathbf{X}_2$ on to a regular grid $\mathbf{G}$. This is equivalent to applying the flow field to itself -- see \cref{eq:apply_op}:
\begin{equation}\label{eq:switch_s}
\begin{split}
    \ft{1}{2} := \left.\fs{1}{2}\right\vert^\text{switch} 
    &= \mathcal{I}\{\Fs{1}{2};\:\mathbf{X}_2 \shortto \mathbf{G}\} \\
    &= \mathcal{I}\{\Fs{1}{2};\:(\mathbf{G} + \Fs{1}{2}) \shortto \mathbf{G}\} \\
    &= \fs{1}{2}\{\Fs{1}{2}\}
\end{split}
\end{equation}

If the optical flow field is currently in the ``target'' reference, the flow vectors need to be interpolated from their start points $\mathbf{X}_1$ on to a regular grid $\mathbf{G}$. Using \cref{eq:flow_inv_ref_switch}, this can again be simplified to a flow application operation:
\begin{equation}\label{eq:switch_t}
\begin{split}
    \fs{1}{2} := \left.\ft{1}{2}\right\vert^\text{switch} 
    &= \mathcal{I}\{\Ft{1}{2};\:\mathbf{X}_1 \shortto \mathbf{G}\} \\
    &= \mathcal{I}\{\Ft{1}{2};\:(\mathbf{G} - \Ft{1}{2}) \shortto \mathbf{G}\} \quad \vert \quad \Ft{1}{2} = -\Fs{2}{1} \\
    &= \mathcal{I}\{\Ft{1}{2};\:(\mathbf{G} + \Fs{2}{1}) \shortto \mathbf{G}\} \\
    &= \fs{2}{1}\{\Ft{1}{2}\}
\end{split}
\end{equation}

\subsection{Inverting Flows}
\label{theory:inv}

If a flow is to be inverted while retaining the same frame of reference, we can chain the operation to invert a flow while switching the reference frame from \cref{eq:flow_inv_ref_switch} and the reference frame switch from \cref{theory:switch}:
\begin{align}
    \fs{2}{1} := \left.\fs{1}{2}\right\vert^\text{inv} 
    &= \left.\ft{2}{1}\right\vert^\text{switch} \label{eq:inv_s} \\
    &= \mathcal{I}\{\Ft{2}{1};\:\mathbf{X}_2 \shortto \mathbf{G}\} \notag \\
    &= \mathcal{I}\{-\Fs{1}{2};\:(\mathbf{G} + \Fs{1}{2}) \shortto \mathbf{G}\} \notag \\ 
    &= \fs{1}{2}\{-\Fs{1}{2}\} \notag \\ \notag \\
    \ft{2}{1} := \left.\ft{1}{2}\right\vert^\text{inv} 
    &= \left.\fs{2}{1}\right\vert^\text{switch} \label{eq:inv_t} \\ 
    &= \mathcal{I}\{\Fs{2}{1};\:\mathbf{X}_1 \shortto \mathbf{G}\} \notag \\
    &= \mathcal{I}\{-\Ft{1}{2};\:(\mathbf{G} - \Ft{1}{2}) \shortto \mathbf{G}\} \quad \vert \quad \Ft{1}{2} = -\Fs{2}{1} \notag \\
    &= \mathcal{I}\{-\Ft{1}{2};\:(\mathbf{G} + \Fs{2}{1}) \shortto \mathbf{G}\} \notag \\
    &= \fs{2}{1}\{-\Ft{1}{2}\} \notag
\end{align}

\subsection{Valid Areas}
\label{theory:valid}

Optical flow fields are unlikely to provide a full mapping between coordinates of the grid $\mathbf{G}$ at time $t_1$ and time $t_2$, resulting in either lost input data, or undefined regions in the warped output data, or both. Consequently, the source domain region whose content is not lost in the application of a specific optical flow field is termed its ``valid source area'', while the ``valid target area'' denotes the target domain region receiving data from the source domain in the warping operation:
\begin{align}\label{eq:valid}
\begin{split}
    \text{valid source of }\f{1}{2} &:= \mathbf{B}(\f{2}{1}\{m\} > 0) \\
    \text{valid target of }\f{1}{2} &:= \mathbf{B}(\f{1}{2}\{m\} > 0)
\end{split}
\end{align}

\noindent where $m$ is a matrix of shape $H \times W$ with the value $1$ everywhere, and $\mathbf{B}()$ is the element-wise evaluation of the given condition on the matrix $m$ warped by the flow $\mathcal{F}$. By definition, $\mathbf{B}()$ yields a boolean array of the same shape $H \times W$ as the input array.

\begin{figure}
    \centering
    \includegraphics[width=0.48\linewidth]{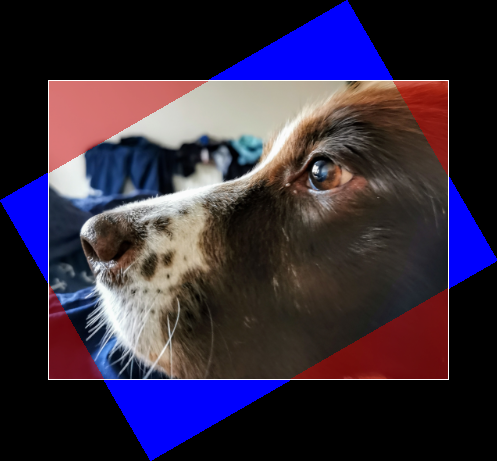}\hfill
    \includegraphics[width=0.48\linewidth]{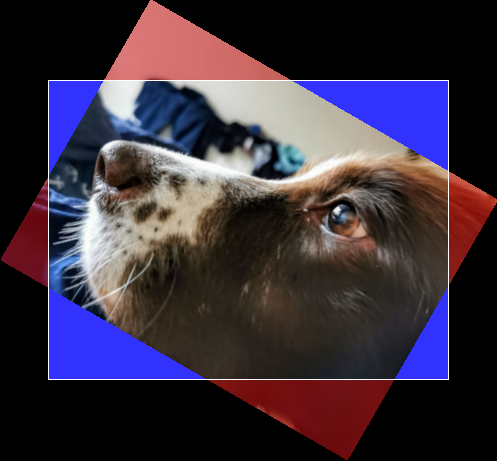}
    \caption{Illustration of the ``invalid area'' concept, using a \SI{30}{\degree} clockwise rotation around the image centre. Red areas denote regions in the source domain lost in the target domain. Blue marks regions in the target domain for which no data can be acquired in the source domain. The white line denotes the image borders, black shows the image padding necessary to avoid any invalid area. \textbf{Left:} The image in the source domain. \textbf{Right:} The rotated image in the target domain.}
    \label{fig:valid_areas}
\end{figure}

The concept is most simply illustrated in \cref{fig:valid_areas}, showing the rotation of a camera relative to an object: new image content unknown at time $t_1$ enters the frame at $t_2$. Given just an image at time $t_1$ and $\f{1}{2}$, the content of this region will remain unknown -- this is termed an ``invalid area'', filled with zero values. It is useful to be able to differentiate this from image regions that also contain the value $0$, but represent known data.

\subsection{Flow composition}
\label{theory:compose}

\begin{figure}
    \centering
    \includegraphics[width=0.9\textwidth]{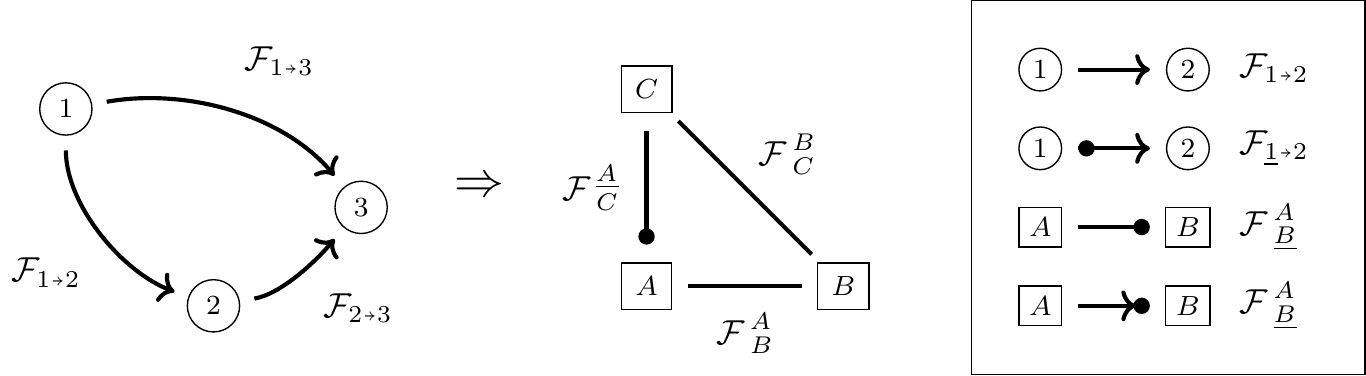}
    \caption{
    \textbf{Left:} Composition of flows $\f{1}{2}$, $\f{2}{3}$, $\f{1}{3}$ between the times $t_1$, $t_2$, $t_3$. Two flows are always known, one is unknown.
    \textbf{Centre:} Generalised representation as flows between the times $A$, $B$, and $C$. $\rf{A}{C}$ is always the unknown (result) flow with its reference $\tref{A}{C}$ at $A$, from which the mapping direction of $\rf{A}{C}$ is inferred. Given this, the two known flows and their directions and references can be matched with $\rf{A}{B}$ and $\rf{B}{C}$.
    \textbf{Right:} Symbol convention. Arrows indicate the temporal mapping direction, dots denote the time $t$ used as reference.
    }
    \label{fig:flow_comb_remapped}
\end{figure}

At the core of \of{} are flow composition operations, defined as follows:
\begin{equation}\label{eq:flow_comp}
    \f{1}{2} \oplus \f{2}{3} = \f{1}{3}
\end{equation}

\noindent where $\oplus$ denotes sequential operations as illustrated on the left in \cref{fig:flow_comb_remapped}. Given any two of the above flow fields, we would like to calculate the third one, resulting in three possible operations, or three ``modes'', whose numbering indicates which flow field is the unknown one. If given in the target frame of reference, the flows in \cref{eq:flow_comp} can be rewritten more explicitly as the following mappings:
\begin{equation}\label{eq:flow_comp_t}
\begin{split}
    (\mathbf{X}_1 \mapsto \mathbf{G}_3) = (\mathbf{X}_1 \mapsto \mathbf{G}_2) \oplus (\mathbf{X}_2 \mapsto \mathbf{G}_3)
\end{split}
\end{equation}

From \cref{eq:flow_comp_t} it is clear that the flow vectors cannot be combined with simple arithmetic operations, as they refer to different coordinate locations at different times. The correct approach is to match mapping locations and times first by performing flow application and interpolation operations, then to add the flow vectors. As an example, for the sequential combination of the flow fields $\ft{1}{2}$ and $\ft{2}{3}$ -- ``mode $3$'' -- we can observe that to add the flow vectors the reference frame of $\ft{1}{2}$ needs to be brought from $t_2$ to $t_3$. It follows:
\begin{equation}\label{eq:flow_comp_t_3}
\begin{split}
    \ft{1}{3} &:= \text{compose}(\ft{1}{2},\:\ft{2}{3}\ \vert\ \text{mode } 3)\\
    &\:\:= \ft{1}{2} \oplus \ft{2}{3} = \Ft{2}{3} + \ft{2}{3}\{\Ft{1}{2}\}
\end{split}
\end{equation}

By remapping the flows to a different representation shown in the centre of \cref{fig:flow_comb_remapped}, this approach can be generalised. The guiding principle is then to get $\rf{A}{B}$ and $\rf{B}{C}$ into the same frame of reference before performing a linear vector addition. If necessary, the result is moved to $t_A$ to ensure $\rf{A}{C}$ is returned in the desired frame of reference. Algorithm \ref{alg:compose} details all steps necessary.

\begin{table}
\normalsize
\centering
\captionsetup{labelformat=empty}   
\caption{\textbf{Algorithm 1}\ General flow composition algorithm}
\label{alg:compose}
\begin{tabular}{@{} w{l}{0.06\textwidth} w{l}{0.93\textwidth} @{}}
    \toprule
    \linenum{1} & \textbf{Define: } $\rftop{A}{C}$ as the unknown flow \\
    \linenum{2} & \textbf{Match: } Reference $A$ of $\rftop{A}{C}$ to the reference of the unknown flow \\
    \linenum{3} & \textbf{Match: } $C$ to the non-reference time domain of the unknown flow \\
    \linenum{4} & \textbf{Match: } $B$ to the remaining time $1$, $2$ or $3$ \\
    \linenum{5} & \textbf{if} $\rf{A}{B}$ from $A$ to $B$ \textbf{then}\\
    \linenum{6} & \qquad \textbf{Define: } $\sign{A}{B} = 1$ \\
    \linenum{7} & \textbf{else if} $\rf{A}{B}$ from $B$ to $A$ \textbf{then} \\
    \linenum{8} & \qquad \textbf{Define: } $\sign{A}{B} = -1$ \\
    \linenum{9} & \textbf{if} $\rf{B}{C}$ from $B$ to $C$ \textbf{then}\\
    \linenum{10} & \qquad \textbf{Define: } $\sign{B}{C} = 1$ \\
    \linenum{11} & \textbf{else if} $\rf{B}{C}$ from $C$ to $B$ \textbf{then} \\
    \linenum{12} & \qquad \textbf{Define: } $\sign{B}{C} = -1$ \\
    \linenum{13} & \textbf{if} $\rf{A}{C}$ from $C$ to $A$ \textbf{then} \hfill $\rhd$ $\rf{A}{C}$ in opposite direction than assumed\\
    \linenum{14} & \qquad $\sign{A}{B}\ \leftarrow\ \sign{A}{B} * -1$\\
    \linenum{15} & \qquad $\sign{B}{C}\ \leftarrow\ \sign{B}{C} * -1$\\
    \linenum{16} & \textbf{if} $\tref{B}{C} = C$ \textbf{then}\\
    \linenum{17} & \qquad $\rf{B}{C}\ \leftarrow\ \rf{B}{C}\vert^\text{switch}$ \hfill $\rhd$ Now $\tref{B}{C} = B$\\
    \linenum{18} & \textbf{if} $\tref{A}{B} = A$ \textbf{and} $\rf{A}{B}$ from $A$ to $B$ \textbf{then}\\
    \linenum{19} & \qquad $\rf{B}{C}\ \leftarrow\ \rf{A}{B}\vert^\text{inv}\{\rf{B}{C}\}$\\
    \linenum{20} & \textbf{else if} $\tref{A}{B} = A$ \textbf{and} $\rf{A}{B}$ from $B$ to $A$ \textbf{then} \\
    \linenum{21} & \qquad $\rf{B}{C}\ \leftarrow\ \rf{A}{B}\{\rf{B}{C}\}$ \hfill $\rhd$ Now $\tref{B}{C} = \tref{A}{B}$, either $A$ or $B$\\
    \linenum{22} & $\rf{A}{C} \leftarrow \sign{A}{B} * \rf{A}{B} + \sign{B}{C} * \rf{B}{C}$ \\
    \linenum{23} & \textbf{if} $\tref{A}{B} = B$ \textbf{and} $\rf{A}{B}$ from $A$ to $B$ \textbf{then}\\
    \linenum{24} & \qquad $\mathcal{F}_{AC}\ \leftarrow\ \mathcal{F}_{AB}\vert^\text{inv}\{\mathcal{F}_{AC}\}$\\
    \linenum{25} & \textbf{else if} $\tref{A}{B} = B$ \textbf{and} $\rf{A}{B}$ from $B$ to $A$ \textbf{then} \\
    \linenum{26} & \qquad $\rf{A}{C}\ \leftarrow\ \rf{A}{B}\{\rf{A}{C}\}$ \hfill $\rhd$ Now $\tref{A}{C} = A$, as required by definition\\
    \noalign{\smallskip}\bottomrule
\end{tabular}
\end{table}
\section{Implementation}

The concrete code implementation of the theoretical underpinnings of \of{} is based on existing functions from Python packages such as PyTorch \cite{pytorch}, NumPy \cite{numpy} and OpenCV \cite{opencv} as far as feasible. This reduces potential error sources and increases performance by making use of highly efficient native functions. \Of{} has been released as two variants: \ofnp{} is based on NumPy arrays, while \ofpt{} uses exclusively PyTorch tensors for internal attribute representations. The latter paves the way for operations to be performed on a GPU, as well as the calculation of gradients ensuring differentiability needed for back-propagation in deep learning algorithms.

Though the core functionality of both variants is the same, \ofpt{} as the more advanced package is the basis for all further discussion. It comprises around $3500$ lines of code as well as comprehensive tests based on the \texttt{unittest} package. The \texttt{coverage} package \cite{coverage} reports an overall test coverage of \SI{99}{\percent}. All code is available open-source under the MIT License. It requires Python $\geq3.7$ and can be installed from the Python Package Index using pip (\verb!pip install oflibpytorch!). Extensive documentation is available online\footnote{https://oflibpytorch.readthedocs.io/}.

\subsection{Structure}
The flow field conceptually laid out in \cref{theory} is implemented as a custom class called \texttt{Flow} with three main attributes:

\begin{itemize}
    \item Vectors \texttt{vecs}: The flow field as a tensor of shape $2 \times H \times W$, following the PyTorch channel-first convention. The dimension of size $2$ corresponds to the vector components $(x, y)$, with x defined positive towards the right, and y defined positive downwards. This follows the OpenCV convention for flow vectors, e.g. as output by the \texttt{calcOpticalFlowFarneback} function \cite{opencv}.
    \item Reference \texttt{ref}: A string \texttt{`s'} or \texttt{`t'} referring to ``source'' or ``target''.
    \item Mask \texttt{mask}: A boolean tensor of shape $H \times W$ indicating valid flow field areas. Any operation on the flow field vectors is also performed on the mask, so valid flow field areas are automatically tracked through all operations.
\end{itemize}

The functions and class methods provided can be grouped by application. \textit{Constructors} allow the user to create flow objects from given flow vectors, any transformation matrix, a list of transforms, or filled with zero-magnitude vectors. Flow \textit{manipulation} includes inverting, resizing, and padding a flow, or switching its frame of reference. Flow \textit{application} refers to tracking points or warping an input. Flow \textit{evaluation} comprises finding the valid source or target area, determining necessary padding values to avoid invalid areas, and fitting a transformation matrix to the flow field. Finally, \textit{visualisation} is achieved by either drawing arrows or using the general convention encoding the vector direction in the image hue and the magnitude as the saturation.

\subsection{Interpolation}

As shown in \cref{theory}, virtually all operations on flow fields are one of two types of interpolations at heart. Interpolating from a regular grid to unstructured points ($\mathbf{G} \shortto \mathbf{X}$) can be easily and accurately achieved with bilinear interpolation. Interpolating from unstructured data onto a grid ($\mathbf{X} \shortto \mathbf{G}$) on the other hand is a much more complex operation. For NumPy arrays, SciPy's \texttt{griddata} function \cite{scipy} offers a solution: it performs a Delaunay triangulation of all input points and uses a k-d tree to find the closest simplex to each unknown value point on $\mathbf{G}$, interpolating between the vertex values using linear barycentric interpolation.

However, empirical measurements show it to be around $400$ times slower than interpolating from a regular grid, an obvious issue for any real-time application. Additonally, \texttt{griddata} cannot be executed on a GPU, is not interoperable with PyTorch tensors, and therefore is also not differentiable in the context of the PyTorch gradient function. Hence, a purely PyTorch-based alternative function was paramount for \ofpt{}. Unfortunately, most of the functions needed for this algorithm are not yet available natively in Pytorch. As a less accurate, but around $100\times$ faster alternative we implemented inverse bilinear interpolation, utilising the principle of allocating unstructured point values to their surrounding grid values, with a weighting inversely proportional to their distance \cite{sanchez_2013_direct,hofinger_2020_improving}. The inverse interpolation function named \verb|grid_from_unstructured_data| is also exposed at module level, as it may well prove useful to other applications.

\Cref{tab:performance} shows a comparison of bilinear interpolation as used for $\ft{1}{2}\{\cdot\}$ (`t') with inverse bilinear interpolation as used for $\fs{1}{2}\{\cdot\}$ (`s'): the latter is still up to $3\times$ slower, but no longer $400\times$ as when \texttt{griddata} is used. The results also show the impact of using a GPU, especially when batched -- if \texttt{griddata} is avoided.

\begin{table}[t]
\centering
\renewcommand{\arraystretch}{1.2}
\setlength{\tabcolsep}{0.1cm}
\caption{Time in \si{\milli\second} for a single flow application operation in both frames of reference (`s', `t'), image size $250\times400\text{\,px}$, on CPU (i$7$-$8750$H) and GPU (GTX $1050$ Ti), averaged over $10000$ tests, or $1000$ tests of batch size $10$ (if indicated).}
\label{tab:performance}
\begin{tabular}{@{} w{l}{1.4cm} w{c}{2.2cm} w{c}{1.5cm} w{c}{1.5cm} w{c}{2.3cm} w{c}{2.3cm}@{}}
    \toprule
     & `s', \texttt{griddata} & `s' & `t' & `s', batch $10$ & `t', batch $10$ \\
    \midrule
    CPU & $1470$ & $33.1$ & $11.9$ & $37.4$ & $9.4$\\
    GPU & n/a & $10.2$ & $6.9$ & $6.4$ & $1.0$\\
    \bottomrule
\end{tabular}
\end{table}

\subsection{Verification}

Empirical verification of the theory behind the \verb|Flow.combine| function as well as its practical implementation was obtained by randomised tests\footnote{Code available from the test folder in the source repository}. In each test, three flow fields of size $150\times250\text{\,px}$ are created: two representing either a rotation, translation, or scaling transform, and one as their sequential application. Rotation and scaling centres are randomly chosen within the flow field area, and motion magnitudes are sampled from a uniform distribution such that the maximum vector magnitude is $50\text{\,px}$. These randomised tests were performed $10000$ times for each flow combination mode, randomising input and output references. The absolute error \E{abs} was calculated as the mean of all flow vector end point errors of the calculated result compared to the true value, and the relative error \E{rel} as the mean of end point errors divided by the true flow magnitude. For both \E{abs} and \E{rel}, only the valid area of the calculated flow field was considered.

\begin{table}[t]
\centering
\renewcommand{\arraystretch}{1.2}
\setlength{\tabcolsep}{0.1cm}
\caption{Percentage of absolute errors \E{abs} as well as relative errors \E{rel} in the result of the \texttt{Flow.combine} function, obtained from $10000$ evaluations for each mode, using random combinations of affine transforms. $N_{\text{vectors}}$ denotes the number of individual vectors used to calculate \E{abs} and \E{rel}.}
\label{tab:accuracy}
\begin{tabular}{@{} w{l}{1cm} w{c}{1.4cm} w{c}{1.4cm} w{c}{1.4cm} w{c}{1.4cm} w{c}{1.4cm} w{c}{1.4cm} w{c}{1.4cm} @{}}
    \toprule
    Mode & $N_{\text{vectors}}$ & \multicolumn{2}{c}{\E{abs} in px} & \multicolumn{2}{c}{\% of \E{abs} below} & \multicolumn{2}{c}{\% of \E{rel} below} \\
    \cmidrule(l{0.2cm}r{0.2cm}){3-4}\cmidrule(l{0.2cm}r{0.2cm}){5-6}\cmidrule(l{0.2cm}){7-8}
    & & mean & max & $0.05\text{\,px}$ & $0.005\text{\,px}$ & $0.005$ & $0.0005$ \\
    \midrule
    1 & \num{3.0e8} & $0.003$ & $0.569$ & $0.995$ & $0.837$ & $0.983$ & $0.816$\\
    2 & \num{3.0e8} & $0.003$ & $0.461$ & $0.995$ & $0.839$ & $0.983$ & $0.820$\\
    3 & \num{3.1e8} & $0.003$ & $0.581$ & $0.996$ & $0.847$ & $0.996$ & $0.894$\\
    \bottomrule
\end{tabular}
\end{table}

Results are listed in \cref{tab:accuracy}. Even given large motions of up to $50\text{\,px}$ present in the optical flow fields, absolute errors remain very small at a mean value of $0.003\text{\,px}$, with over \SI{99}{\percent} below $0.05\text{\,px}$ and the maximum around $0.5\text{\,px}$. The relative error \E{rel} is under \SI{0.5}{\percent} for a vast majority of the flow vectors.




\subsection{Application}

Training deep learning algorithms requires large amounts of annotated data. One way of obtaining this is the generation of synthetic data. In the context of optical flow prediction, this means an image pair with an associated ground truth optical flow that encodes the motion from one image to the other. \Of{} can be used to generate this efficiently: given a base image $1$ and two flows $\f{1}{2}$ and $\f{1}{3}$ used to arrive at the images $2$ and $3$, $\f{2}{3}$ as the ground truth flow between this image pair is easily obtained by the use of the \texttt{Flow.combine(mode=2)} function.

If done naively, the intrinsic nature of these composition operations will leave invalid areas in the resulting flow field, containing incorrect flow information -- obviously to be avoided for a ground truth. The solution is to apply padding to the flow fields before using \texttt{Flow.combine}. We can not only determine the right amount of padding using \texttt{Flow.get\_padding}, we can also deliberately choose the frames of reference to simplify this operation: If $\f{2}{3}$ is required in the target frame of reference, then drawing it in the generalised $ABC$ representation from \cref{fig:flow_comb_remapped} and applying algorithm $1$ suggests using $\fs{1}{2}$ (``source'') and $\ft{1}{3}$ (``target''). This choice ensures the only operation outside of the (linear) vector addition is $\ft{1}{3}\{\Fs{1}{2}\}$, i.e. the application of $\ft{1}{3}$ to $\fs{1}{2}$. This means the required padding can be directly calculated from $\ft{1}{3}$.

If, say, the source-based flow field $\fs{1}{2}$ itself was the result of a composition of several individual motion effects, again the use of padding needs to be considered. Using the same considerations as before we determine the source frame of reference to be most convenient for the individual motions making up $\fs{1}{2}$, with the required padding again easily obtainable with \texttt{Flow.get\_padding}.

\begin{figure}
    \centering
    \includegraphics[width=1\textwidth]{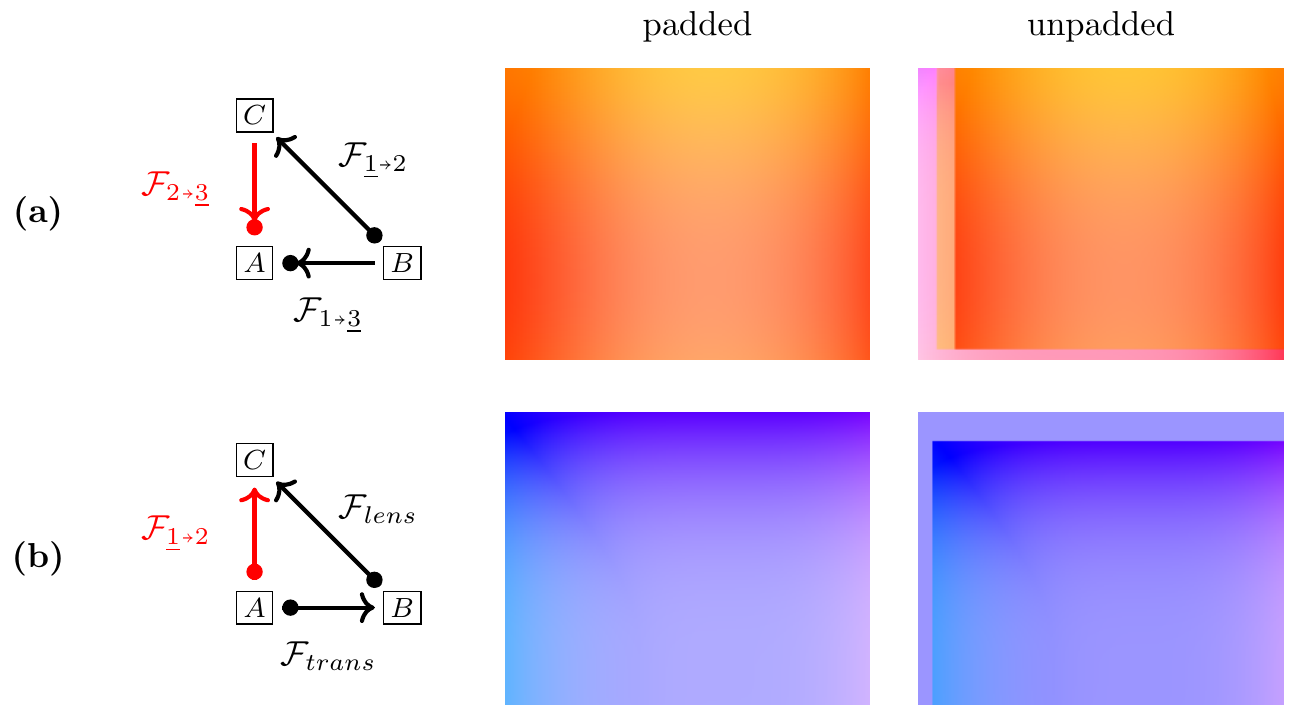}
    \caption{An illustration of the effect of using padding, with flow fields visualised on the right-hand side.
    \textbf{(a)} The $ABC$ representation of the operation that combines $\ft{1}{3}$ and $\fs{1}{2}$ to calculate $\ft{2}{3}$, and a visualisation of the result. 
    \textbf{(b)} The $ABC$ representation of the operation that combines the translation $\mathcal{F}_{trans}$ and the lens distortion $\mathcal{F}_{lens}$ to obtain $\fs{1}{2}$ from (a), all in the source frame of reference.
    }
    \label{fig:usage}
\end{figure}

A working example is provided in the code snippet below. $\f{1}{2}$ and $\f{1}{3}$ each consist of a translation combined with a lens warping effect simulated by a third-order polynomial function of the distance from the lens centre, obtained with the help of the scaling transform built into \of{}. This could simulate a medical imaging scenario, in which a tissue is seen through a moving lens while itself being shifted relative to the camera. \Cref{fig:usage} shows the $ABC$ representation of the flows, along with the effect of accounting for the required padding.

\begin{verbatim}
size = (200, 250)
lens1 = [['scaling', 110, 120, 1.02]]
lens2 = [['scaling', 140, 160, 1.02]]
trans1 = [['translation', 20, -10]]
trans2 = [['translation', -10, -20]]

# Create flow field F13
flow_lens = Flow.from_transforms(lens1, size, 't')
flow_lens.vecs **= 3
flow_trans = Flow.from_transforms(trans1, size, 't')
f13 = flow_trans.combine(flow_lens, 3)

# Create flow field F12
pad1 = f13.get_padding(0)                   # Padding for f23
flow_trans = Flow.from_transforms(trans2, size, 's', padding=pad1)
pad2 = flow_trans.get_padding(0)            # Padding for f12
pad3 = [sum(p) for p in zip(pad1, pad2)]    # Total padding
flow_lens = Flow.from_transforms(lens2, size, 's', padding=pad3)
flow_lens.vecs **= 3
f12 = flow_trans.pad(pad2).combine(flow_lens, 3).unpad(pad2)
# Calculate flow field F23
f23 = f12.combine(f13.pad(pad1), 2, 't').unpad(pad1)
\end{verbatim}

The above example is a simplified version of what lies at the heart of a complex and layered algorithm developed by Ravasio et al. \cite{ravasio_2020_learned} to create synthetic images and ground truth optical flows. These were used successfully to train an optical flow estimation model aimed at tracking points of interest on the retina.
\section{Conclusion}

This work presents a coherent and structured framework for the handling and manipulation of optical flow fields, implemented as a Python library \of{} in two variants, \ofnp{} and \ofpt{}. These packages allow for a multitude of operations on optical flow fields, the theoretical basis of which is derived from first principles. Extensive unit tests as well as empirical validation serve to verify the consistency and accuracy of the formulas derived. 

The flow field composition algorithm developed is shown to be of use for applications such as the creation of large synthetic training datasets for the optical flow prediction algorithm employed by Ravasio et al. \cite{ravasio_2020_learned}. It was also demonstrated that the frame of reference can be chosen specifically to optimise the operations, in this case simplifying the calculation of necessary flow padding.

\Ofpt{} is fully differentiable in the context of the PyTorch gradient function, and therefore ready to be employed in deep learning algorithms that require the use of back-propagation. The code is open-source and easily installed via \verb|pip| (\verb|pip install oflibpytorch|), and we expect it to be of good use to both researchers and non-academic users alike.
\subsection*{Acknowledgements}

This work was supported by the National Institute for Health Research NIHR (Invention for Innovation, i4i; NIHR$202879$). The views expressed are those of the authors and not necessarily those of the NHS, the NIHR, or the Department of Health and Social Care.

\clearpage
%
%
\bibliographystyle{splncs04}
\bibliography{bibliography}
\end{document}